OPINION

# From artificial to organic: Rethinking the roots of intelligence for digital health

Prajwal Ghimire[1,2]*, Keyoumars Ashkan[1,2,3]

1 School of Biomedical Engineering & Imaging Sciences, King's College London, London, United Kingdom, 2 Department of Neurosurgery, King's College Hospital NHS Foundation Trust, London, United Kingdom, 3 Institute of Psychology, Psychiatry and Neuroscience, King's College London, London, United Kingdom

* prajwal.1.ghimire@kcl.ac.uk

## Abstract

The term "artificial" implies an inherent dichotomy from the natural or organic. However, AI, as we know it, is a product of organic ingenuity—designed, implemented, and iteratively improved by human cognition. The very principles that underpin AI systems, from neural networks to decision-making algorithms, are inspired by the organic intelligence embedded in human neurobiology and evolutionary processes. The path from "organic" to "artificial" intelligence in digital health is neither mystical nor merely a matter of parameter count—it is fundamentally about organization and adaption. Thus, the boundaries between "artificial" and "organic" are far less distinct than the nomenclature suggests.

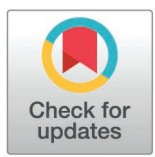

## Introduction

The mid-20th century was a formative era for the study of machine intelligence. In 1950, the British mathematician Alan Turing proposed a thought experiment—later known as the Turing Test—to probe a fundamental question: could a machine ever think? Turing argued that if a computer could execute a conversation so seamlessly that a human judge could not distinguish it from a real person, then, for all practical purposes, the machine was "thinking" [1]. His idea gave early researchers a criterion for comparing artificial behavior to human cognition, even if no one believed it to be a perfect or final measure.

Just a few years later, in 1956, the Dartmouth Summer Research Project on Artificial Intelligence brought together a small group of visionary scientists [2]. They gave the new field its name, Artificial Intelligence (AI), and set forth the bold goal of replicating or exceeding human cognitive capabilities in non-biological substrates. These early pioneers approached their work as a grand quest to construct minds out of silicon and algorithms, rather than flesh and neurons. If Turing's thought experiment was a philosophical spark, the Dartmouth gathering ignited an entire academic discipline.

From these beginnings, the notion took hold that machine intelligence might evolve into a distinct and separate entity, growing ever more sophisticated until it approached or even surpassed human intellect [3].

**Roots: Human inputs and patterns of thought**

To date, we continue to use Turing's framework and the Dartmouth-inspired term "Artificial Intelligence". Yet as AI technology has advanced, and particularly as data-driven machine learning systems have come to dominate the field, our understanding of what makes these systems "intelligent" has shifted. Instead of observing entirely new forms of reasoning emerging from isolated digital minds, we see something more nuanced: these systems are deeply and inescapably rooted in human inputs, human culture, and human patterns of thought [4].

For all the complexity of modern machine learning, the fact remains that today's AI models learn from data we provide. Whether they are identifying objects in images, translating languages, recognizing speech, or engaging in human-like conversation, their abilities flow from patterns observed in massive, human-curated datasets [5]. The clever turns of phrase in a language model's output are echoes of human writing. The refined decision-making of a recommendation system arises from signals in human behavior. Even the architecture of neural networks are designed, tuned, and improved by people drawing inspiration from biological brains and mathematical insights [6].

**Terminology: Artificial, organic, and intelligence**

This interconnectedness underscores a crucial point: what we call "artificial" intelligence is not, in reality, conjured out of a void. Instead, it is a distillation of our collective intelligence, channeled and rearranged by algorithms. The very term "artificial" might suggest that these systems are something other than human in their origins, but this framing can be misleading. Just as languages evolve through communities of speakers, and just as cultural knowledge passes through generations of human minds, the "intelligence" in AI emerges from the human intellectual ecosystem it was trained on. Machines do not intrinsically know how to parse a sentence or evaluate the correctness of a fact—these capabilities only arise from exposure to our texts, images, and examples.

This realization changes how we understand Turing's challenge and the Dartmouth vision. When an AI passes a Turing-like test of conversational skill, it is not proving that it possesses some newly minted, inorganic mind. Rather, it is demonstrating how adeptly it can replicate human conversational patterns. If we cannot tell whether the speaker behind the screen is human or machine, it is because the machine is reflecting, in refined statistical form, the human-made patterns that taught it how to speak in the first place. The "intelligence" we see is, at its core, a reflection of the organic intellect that produced the input data [7].

Acknowledging this human backbone to machine intelligence also carries implications for ethics, accountability, and design. If what we label as "artificial" is in fact "organically" derived—from human knowledge, human choices, and human

PLOS Digital Healthbiases—then we remain responsible for the outcomes. If a biased dataset led to a biased model, then the root cause is not some alien mind, but our own flawed inputs [8]. Understanding AI as organic in its essence encourages us to scrutinize the data we feed it and the purposes we set. It reminds us that the machine's "values" are, in truth, our values writ large and automated at scale.

Thus, we define organic intelligence not by the material of its substrate but by organization of its dynamics: systems that exhibit self-organization, adaptive plasticity, and hierarchical feedback control. In this view, the so-called artificial architectures of deep learning are themselves extension of organic principles materialized through inorganic means.

## Organically-rooted constructs

This perspective does not diminish the genuine achievements of AI research. On the contrary, it highlights an extraordinary human accomplishment to, in essence, extend this capability such that the intelligence, previously confined to the realm of organic matter, can now also be delivered by the inorganic matter. We have thus created tools that can amplify, reorganize, and reflect our collective intellect in powerful new ways. These systems have the capacity to make certain tasks easier, to unearth patterns we might have missed, and to serve as creative partners in fields from science to the arts. By seeing them for what they are, organically-rooted human-guided constructs, we can better integrate them into society, ensuring they complement rather than distort our priorities [9].

## Concept of artificial general intelligence, superintelligence, and digital health

Recent progress has provided a base for development of artificial general intelligence (AGI) and ultimately super general intelligence (SGI) [10–14]. This has been possible due to enhancements of hardware capabilities. The core idea still is to try to mimic the human brain networks to achieve skills such as multitasking, reasoning, and establishing causal relationships with multiple predictions, but not without inherent bias that comes from the training performed by human users [10–14]. These models are likely to be useful for digitizing hospital-related tasks, enhancing the experience of digital health. This will be possible only if speed is balanced with accountability; explainability is treated as part of safety and targeting human-level breadth across tasks under resource constraints. In clinical settings, these aspects translate into uncertainty-aware objectives, rehearsed rollback protocols, and escalation pathways [10,14].

Some examples of AI applications in healthcare with organic roots would be sparse and modular architectures for radiology triage and neuro-oncology stratification, which align with small-world efficiency principles [11,12,15]. Other examples include continual and domain-adaptive learning models with homeostatic calibration for cross-site generalization [4,5] along with hybrid neuro-symbolic and memory augmented networks that integrate reasoning and perception for longitudinal patient monitoring [16]. The main technical barriers for applying these models would be data quality and harmonization across institutions, generalization and calibration under scanner protocol drift, compute and energy constraints for edge deployment, and governance of adaptive models [17].

The potential milestones that will be required before AGI and SGI can be considered for healthcare contexts could be robust narrow AI (modular clinical models with drift detection with safeguards including dataset harmonization and calibration metrics); cross-task generalization (unified triage, segmentation and report generation with safeguards including explainability and abstention frameworks); tool-use and reasoning (integration with EHR and external databases with safeguards of continuous auditing and model cards) and autonomy under oversight (context-aware multi-agent reasoning with safeguards including uncertainty-over-preference, corrigibility and off-switch verification) [18–22]. These milestones provide a bridge from current digital health AI to the AGI/SGI discourse.

Bias mitigation can be achieved within algorithmic design through re-weighting and counterfactual data augmentation during training, combined with structural plasticity that can down-weight spurious or site-specific connections over time [23–25]. On the other hand, accountability can be achieved by embedding governance hooks such as logging rewiring events, explanation stability, or abstention triggers directly into model architecture, ensuring traceable model evolution

PLOS Digital Health | https://doi.org/10.1371/journal.pdig.0001109    December 1, 2025    3 / 5

[26]. This leads to bias mitigation and accountability engineered into system's logic rather than being deferred to clinical workflow.

## Implications of shift in terminology

The shift in terminology from real/artificial to organic/inorganic has implications for research priorities, benchmark design, and interdisciplinary collaboration [12]. It can thus lead to a paradigm change in research focus from model scale to organizational efficiency; proposing dynamic benchmarks that test adaptability and calibration under distribution shift, rather than static accuracy-only leaderboards. Furthermore, the organic/inorganic lens invites joint design between neuroscientists, clinicians, and AI engineers—treating intelligence as a continuum of structure and adaptation rather than a categorical divide [12,13,27–29]. These shifts will reframe AI research as an integration science, aligning computational modeling with biological organization and cognitive safety.

## Conclusion

In the decades since Turing's and the Dartmouth pioneers' era, we have advanced toward systems that can meet and sometimes surpass the benchmarks of that era's imagination. But we have also learned that "artificial" intelligence cannot be neatly separated from the human context that birthed it [30]. The name may endure out of historical convenience, but as we chart the future of AI with superintelligence, it may be more accurate to think of these technologies as inorganic channels for organic wisdom, extended and transformed through computational means. The time, perhaps, is now right for a name rethink away from real versus artificial intelligence towards organic versus inorganic intelligence as we make our move towards digital health in our hospitals. After all, intelligence, organic or inorganic, is defined by how systems organize and adapt information.

## Author contributions

**Conceptualization:** Prajwal Ghimire, Keyoumars Ashkan.

**Investigation:** Prajwal Ghimire.

**Methodology:** Prajwal Ghimire.

**Project administration:** Keyoumars Ashkan.

**Resources:** Prajwal Ghimire.

**Writing – original draft:** Prajwal Ghimire.

**Writing – review & editing:** Prajwal Ghimire, Keyoumars Ashkan.